%% file: root.tex
\documentclass[letterpaper, 10 pt, conference]{ieeeconf}  % Comment this line out if you need a4paper

\IEEEoverridecommandlockouts                              % This command is only needed if 
\usepackage{graphicx} % for pdf, bitmapped graphics files
\usepackage{epsfig} % for postscript graphics files
\usepackage{mathptmx} % assumes new font selection scheme installed
\usepackage{times} % assumes new font selection scheme installed
\usepackage{amsmath} % assumes amsmath package installed
\usepackage{amssymb}  % assumes amsmath package installed
\usepackage{booktabs}
\usepackage{multirow}
\usepackage{caption}
\usepackage{subcaption}
\usepackage[table]{xcolor}% http://ctan.org/pkg/xcolor
\usepackage{xcolor}
\usepackage{calc}

\usepackage{hyperref}  % for links

\graphicspath{{./graphics/}}

\renewcommand{\H}{$\mathbb{H}$}
\newcommand{\I}{$\mathbb{I}$}

\newcommand{\car}{\textit{car}}
\newcommand{\bicycle}{\textit{bicycle}}
\newcommand{\motorcycle}{\textit{motorcycle}}
\newcommand{\truck}{\textit{truck}}
\newcommand{\othervehicle}{\textit{other vehicle}}
\newcommand{\person}{\textit{person}}
\newcommand{\motorcyclist}{\textit{motorcyclist}}
\newcommand{\bicyclist}{\textit{bicyclist}}
\newcommand{\road}{\textit{road}}
\newcommand{\parking}{\textit{parking}}
\newcommand{\sidewalk}{\textit{sidewalk}}
\newcommand{\otherground}{\textit{other ground}}
\newcommand{\building}{\textit{building}}
\newcommand{\fence}{\textit{fence}}
\newcommand{\vegetation}{\textit{vegetation}}
\newcommand{\trunk}{\textit{trunk}}
\newcommand{\terrain}{\textit{terrain}}
\newcommand{\pole}{\textit{pole}}
\newcommand{\trafficsign}{\textit{traffic sign}}
\definecolor{car}{RGB}{100, 150, 245}
\definecolor{bicycle}{RGB}{100, 230, 245}
\definecolor{motorcycle}{RGB}{30, 60, 150}
\definecolor{truck}{RGB}{180, 30, 80}
\definecolor{othervehicle}{RGB}{0, 0, 255}
\definecolor{person}{RGB}{255, 30, 30}
\definecolor{motorcyclist}{RGB}{150, 30, 90}
\definecolor{bicyclist}{RGB}{255, 40, 200}
\definecolor{road}{RGB}{255, 0, 255}
\definecolor{parking}{RGB}{255, 150, 255}
\definecolor{sidewalk}{RGB}{75, 0, 75}
\definecolor{otherground}{RGB}{175, 0, 75}
\definecolor{building}{RGB}{255, 200, 0}
\definecolor{fence}{RGB}{255, 120, 50}
\definecolor{vegetation}{RGB}{0, 175, 0}
\definecolor{trunk}{RGB}{135, 60, 0}
\definecolor{terrain}{RGB}{150, 240, 80}
\definecolor{pole}{RGB}{255, 240, 150}
\definecolor{trafficsign}{RGB}{255, 0, 0}
\definecolor{low}{RGB}{0, 35, 80}
\definecolor{high}{RGB}{255, 230, 50}
\newlength{\DepthReference}
\newlength{\HeightReference}
\newlength{\Width}%
\newcommand{\textb}[1]%
{%
	\settowidth{\Width}{#1}%
	\colorbox{#1}%
	{%      
		\raisebox{-\DepthReference}%
		{%
			\parbox[b][\HeightReference+\DepthReference][c]{\Width}{\centering#1}%
		}%
	}%
}
\newcommand{\textd}[1]{\textcolor{#1}{#1}}

\title{\LARGE \bf
	On the Calibration of Uncertainty Estimation in LiDAR-based Semantic 
	Segmentation}
\author{Mariella Dreissig$^{1, 2, *}$, Florian Piewak$^{1}$ and Joschka Boedecker$^{2}$% <-this % stops a space
\thanks{Acknowledgement: This publication was compiled as part of the research 
project "KI Delta Learning" (project number: 19A19013A) funded by the Federal 
Ministry for Economic Affairs and Energy (BMWi) based on a resolution of the 
German Bundestag.}% <-this % stops a space
\thanks{$^{1}$Mercedes-Benz AG, $^{2}$University of Freiburg, $^{*}$Primary 
contact: {\tt\small mariella.dreissig@mercedes-benz.com}}%
}

\begin{document}

\maketitle
\thispagestyle{empty}
\pagestyle{empty}

%%%%%%%%%%%%%%%%%%%%%%%%%%%%%%%%%%%%%%%%%%%%%%%%%%%%%%%%%%%%%%%%%%%%%%%%%%%%%%%%
\begin{abstract}
	The confidence calibration of deep learning-based perception models plays a crucial role in their reliability. Especially in the context of autonomous driving, downstream tasks like prediction and planning depend on accurate confidence estimates. In the point-wise multiclass classification tasks of sematic segmentation the model naturally has to deal with heavy class imbalances. Due to their underrepresentation, the confidence calibration of classes with smaller instances is challenging but essential, especially for safety reasons. We propose a metric to measure the confidence calibration quality of a semantic segmentation model with respect to individual classes. It utilizes the computation of sparsification curves based on uncertainty estimates. With the help of this classification calibration metric, uncertainty estimation methods can be evaluated with respect to their confidence calibration of any given classes. We furthermore suggest a method to automatically find label problems using this metric to improve the quality of hand- or auto-annotated datasets.
\end{abstract}

%%%%%%%%%%%%%%%%%%%%%%%%%%%%%%%%%%%%%%%%%%%%%%%%%%%%%%%%%%%%%%%%%%%%%%%%%%%%%%%%
\input{sections/introduction}

\input{sections/literature}
\input{sections/methods}
\input{sections/results}
\input{sections/discussion}

\input{sections/conclusion}

% bibliography
\bibliographystyle{IEEEtran}
\bibliography{IEEEabrv,root}

\end{document}

%% file: sections/introduction.tex
\section{Introduction}
\label{sec:introduction}

Environment perception is a prerequisite for most autonomous driving systems. 
It allows the vehicle to detect and understand the behavior of other 
participants and enables it to adapt its own behavior accordingly. Deep 
learning methods took performance in evironment perception to a new level by 
constructing models which can evaluate large amounts of data. These data are 
gathered by various sensors with different modalities, such as cameras, radar 
and LiDAR sensors. In the past few years, the LiDAR sensor gained relevance in 
the context of environment perception for autonomous vehicles due to the added 
value of highly precise depth information.

Besides perception tasks like object detection and tracking, semantic 
segmentation plays a crucial role in scene understanding for autonomous 
vehicles. The task of a semantic segmentation model is to classify every 
measurement individually to obtain a detailed representation of the environment 
including e.g. drivable space, vulnerable road users or non-movable objects. 
The point-wise multi-class semantic segmentation of semi-dense LiDAR data based 
on deep learning has been studied thoroughly during the last years 
\cite{Piewak2020}, \cite{He2021}. Despite major advances in this field, the 
task of semantic segmentation comes with the challenge of handling severely 
imbalanced data \cite{Behley2021}. This is due to the natural distribution of 
spaces and objects, i.e. in a traffic scene there always will be significantly 
more measurements of the road or buildings than of persons or bicyclists. 

\begin{figure}[ht]
	\centering
	\includegraphics[width=0.9\columnwidth]{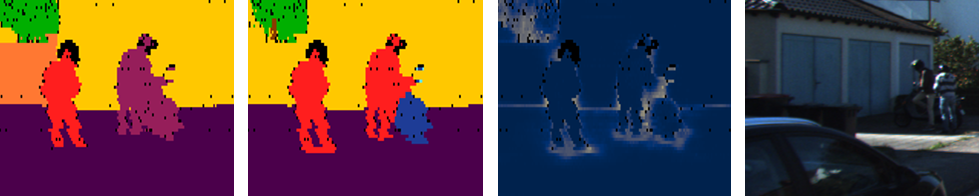} \\
	{\scriptsize semantic segmentation: \textd{vegetation} \textd{trunk} \textb{building} \textd{fence} \textd{person} \textd{motorcyclist} \textd{motorcycle} \textb{bicycle} \\
	uncertainty estimation: \textd{low}, \textb{high}}\\
	\caption{Example of the semantic segmentation of a \person{} and a \motorcyclist{} captured in a projected LiDAR-scan. From left to right: ground-truth label, model prediction, estimated predictive uncertainty. In this case, the model does not predict the label \motorcyclist{} but instead both \person{} and \motorcycle{} with a high confidence. This is technically not incorrect and renders the labeling questionable in this case.}
	\label{fig:example_motorcyclist}
\end{figure}

Althought the current state-of-the-art methods reach a high performance, a deeper understanding of underrepresented classes within the model weights is still challenging. In addition, although deep learning is a powerful tool and successful at generalizing within a known domain, it is prone to failure in situations they have rarely encountered before. Uncertainty estimation can help to address both of these challenges since it allows the model to to detect its own boundaries by providing confidence estimations with their predictions or classifications \footnote{In this work, uncertainty and confidence are used as terms for the same concept. They are defined as complementary to each other: $\text{confidence} = 1 - \text{uncertainty}$.} and it increases the transparency of the classification.
Furthermore, in the context of LiDAR-based computer vision it is especially important to harvest all available information, due to the natural sparsity of the data. Uncertainty estimation helps to gather additional insights of the learning process, especially in the case of imbalanced data.

Downstream tasks rely on well calibrated confidence estimates, which match the actual performance. This allows the sensor fusion or behavior planning to reliably interpret 
the models abilities. In the past, the softmax probabilities were sometimes interpreted as the model confidence, which leads to overconfident estimates \cite{Guo2017}. This 
indicates the need of alternative uncertainty estimation techniques as well as for suitable evaluation metrics to assess their performance.

There are a number of well established methods to estimate uncertainty in deep 
learning models \cite{Gal2016}, \cite{Lakshminarayanan2017}, 
\cite{Kendall2017}, \cite{Arnez2020}. These approaches describe different takes 
on measuring data and model uncertainty respectively, indicating that the exact 
definition and measurement of the different uncertainty types is still an 
ongoing research topic \cite{Mukhoti2021b}.

To make a step towards understanding the confidence calibration of underrepresented classes in semantic segmentation tasks, we elaborate in this paper how the different uncertainty estimation methods behave under different training conditions. For that, we propose a metric to appropriately capture the confidence calibration of a semantic segmentation model and demonstrate its relevance with a series of experiments. Our proposed metric can furthermore be used to uncover questionable labels (compare Fig. \ref{fig:example_motorcyclist}) and paves the path for a qualitative inspection of potentially insufficient label representation. 

The contributions of this paper can be concluded as:
\begin{itemize}
	\item a flexible metric to assess the uncertainty estimation abilities of semantic segmentation models 
	\item an evaluation of underrepresented classes in the context of LiDAR-based semantic segmentation
	\item a novel method to detect label problems based on calibrated deep learning models 
\end{itemize}

%% file: sections/literature.tex
\section{Related Work}
\label{sec:literature}
The following chapter introduces existing literature related to the uncertainty estimation and its assessment in classification tasks like semantic segmentation.

\subsection{Semantic Segmentation of LiDAR point clouds}
In the last years, various approaches on the semantic segmentation of point clouds have been proposed. Some operate in 3D space by utilizing voxels \cite{Huang2016}, \cite{Meng2018} or unordered point clouds \cite{Qi2017}, \cite{Li2018}, \cite{Yang2019}.
Other approaches project the point cloud into the 2D space in order use Convolutional Neural Networks for the semantic segmentation of the range view \cite{Miloto2019}, \cite{Aksoy2019}, \cite{Wu2017}, \cite{Xu2020}. 

Class imbalance is a problem which needs to be adressed in semantic segmentation. The well-established mean Intersection over Union (IoU) evaluation metric accounts for this issue and ensures that underrepresented classes are taken into account appropriately. The metric is widely used to assess the performance of a semantic segmentation model. Common ways to deal with that imbalance is to weight the loss function in favor of underrepresented classes \cite{Paszke2016}, \cite{Miloto2019} or to construct an architecture which is able to overcome the issues imposed by smaller instances \cite{Piewak2020}, \cite{Xu2021}, \cite{Cheng2021}.

\subsection{Uncertainty Estimation in Classification Tasks}
Generally speaking, there are two different kinds of uncertainty: the epistemic (model) uncertainty and the aleatoric (data) uncertainty \cite{Huellermeier2021}. The model uncertainty reflects the uncertainty about the model's own learned parameters, which is due to lack of data, underrepresented effects or hidden information. The data uncertainty reflects the statistical uncertainty, which is due to noise inherent in all processes and effects. Both uncertainties combined compose the overall predictive uncertainty.

In the past years, several approaches on estimating the different uncertainty 
types emerged. Two popular well-established ones are Monte-Carlo Dropout (MCD) 
\cite{Gal2015}, \cite{Gal2017}, \cite{Kamath2020} and deep ensembles 
\cite{Lakshminarayanan2017} \cite{Liu2019}, \cite{Cygert2021}. Both followed 
the idea of Bayesian deep learning and proposed methods which allow the models 
to infer probability distributions over the available classes. Thus, the model 
approximates the true posterior of the data distribution. Yarin Gal 
\cite{Gal2016} stated that the model uncertainty then can be extracted from the 
Monte-Carlo samples by calculating the Mutual Information between the 
prediction and the model parameters posterior. The predictive uncertainty is 
indicated by the entropy over all classes. 

Complementary to that, Kendall et al. \cite{Kendall2017} introduced a technique 
to model uncertainty estimation during training by sampling from the logits 
before applying the softmax function for classification. Thereby they 
demonstrated that the data uncertainty can be learned directly from the input, 
since it can be seen as a function of the data itself. This helps the model to 
measure and deal with data uncertainty by introducing a certain robustness into 
the classification.

\subsection{Evaluation of Uncertainty Estimations} \label{subsec:u_metric}
Guo et al. \cite{Guo2017} proposed the Expected Calibration Error (ECE) to 
assess the confidence calibration of a given neural network. It is defined as the 
expectation between the confidence and the accuracy of the model and is 
calculated by binning the confidence values, computing the according accuracies 
and taking the weighted average of the difference between the accuracy and the 
confidence. While this works well for tasks which actually use the accuracy as 
a performance metric, this approach tends to overestimate the performance for 
underrepresented classes \cite{Piewak2020}. The authors of \cite{Nixon2019} 
adapt this approach to multiclass settings, which in theory make it possible to 
calculate this metric for semantic segmentation tasks and weighting out the 
class-imbalances in the final score. Nevertheless, the previously described 
problem of an overestimated performance still persists. 

Mukhoti et al. \cite{Mukhoti2021a} suggested a metric to capture particular 
parts of the confidence calibration abilities of a semantic segmentation model. Their 
developed technique divides each frame into single patches, from which a 
confusion matrix is constructed. It contains the number of patches which are 
accurate and certain, accurate and uncertain, inaccurate and certain and 
inaccuracte and uncertain. The  conditional probabilities 
$p(\text{accurate}\vert\text{certain})$, 
$p(\text{uncertain}\vert\text{inaccurate})$ and their combination Patch 
Accuracy vs. Patch Uncertainty (PAvPU) can then be calculated. These metrics 
are able to catch two properties:
\begin{itemize}
	\item accuracy in case of confident predictions
	\item uncertainty in case of inaccurate predictions
\end{itemize}
Nevertheless, with respect to subsequent algorithms which rely on well calibrated outputs of a semantic segmentation model, this should be extended to:
\begin{itemize}
	\item confidence in case of accurate predictions
\end{itemize}
Ideally, if it is possible to train the model well enough to reduce the 
epistemic uncertainty as far as possible so that e.g. the entropy only reflects 
the actual data uncertainty, a confidence calibration evaluation metric should also be 
able to capture this effect. Our proposed method has this desired property.

\cite{Ilg2018} proposes a confidence calibration metric for optical flow models. It is derived from Sparsification Plots, which have previously been used to evaluate uncertainty estimates. Through its point-wise nature it can easily be adapted to other point-wise perception tasks, like semantic segmentation. The authors of \cite{Gustafsson2020} have used it in combination with the Brier Score to determine the confidence calibration performance of semantic segmentation models. Yet, the inherent problem with the Brier Score is that it punishes deviations from the one-hot encoding, which contradicts the aim to combat overconfidence in classification models.

%% file: sections/methods.tex
\section{Methods}
\label{sec:methods}
The trainings presented in the following section were conducted on the training sequences of SemanticKITTI dataset \cite{Behley2019}, projected into the 2D sensor view representation \cite{Triess2020}. 

\subsection{Semantic Segmentation Model and Training} \label{subsec:training}
To train a semantic segmentation model, a simplified version of the SalsaNext architecture is used \cite{Cortinhal2020}. It should be noted here that any semantic segmentation backbone will work in this context, especially those which come with an improved performance on imbalanced data \cite{Piewak2019}. 

To evaluate different training strategies, we trained our model once with the uncertainty-aware loss function for classification tasks as proposed in \cite{Kendall2017} and once we extended it to a weighted cross-entropy loss function to account for class imbalances \cite{Paszke2016}. The weights for each class are calculated according to their percentage of points $f_c$ compared to the total points:
\begin{equation}
	weight_{c} = \frac{1}{\text{log}(1.02 + f_c)}
\end{equation}
Following this equation (taken from \cite{Paszke2016}), underrepresented 
classes like \motorcyclist{} receive higher weights than well represented 
classes, e.g \road{}. For further analysis on the distribution of data-points and -labels in the SemanticKITTI dataset, see \cite{Triess2019}.

\subsection{Uncertainty Estimation Technique}
To capture the data and the model uncertainty, both the MCD \cite{Gal2016} (30 samples) and logit sampling (10 samples) \cite{Kendall2017} are used. 
In sum three uncertainty estimation methods are evaluated:
\begin{itemize}
	\item learned aleatoric uncertainty (cf. Equation \ref{eq:logit_sampling})
	\item MCD-based model uncertainty \I \, (cf. Equation \ref{eq:mutual_info})
	\item MCD-based predictive uncertainty \H \, (cf. Equation \ref{eq:entropy})
\end{itemize}

The aleatoric uncertainty estimation is directly incorporated into the training process by implementing a mean and standard deviation for every class instead of one single logit score:
\begin{align} \label{eq:logit_sampling}
	\begin{split}
		x_{i,s} & = f_i^{w} + \sigma_i^w \epsilon_t, \quad \epsilon_s \sim \mathcal{N}(0,I)\\
		Loss_x & = \sum_i \text{log} \frac{1}{S} \sum_s \exp (x_{i, s, c} - \text{log} \sum_{c'} \exp x_{i, s, c'})
	\end{split}
\end{align}
with $i$ denoting the data points, $S$ the logit samples and $c$ the classes. 
The logit vector $x_i, s$ is sampled from a Gaussian distribution defined 
through the models outputs $f_i^w$ and $\sigma_i^w$ given the current mode 
weights $w$ and input frame $x \in D_{train}$. The model samples 10 times from 
these Gaussian distributions and then outputs the class with the highest mean 
softmax score. This softmax probability is proven to be more calibrated than 
the regular softmax \cite{Kendall2017}.

For the MCD process the model is trained with dropout. At test time, multiple forward passes are conducted with dropout enabled, which imitates approximating the posterior over the model weights. The Shannon Entropy (\H) over the class scores then denotes the predictive uncertainty:
\begin{multline} \label{eq:entropy}
	\mathbb{H}[y \vert x, D_{train}] = \\
	- \sum_c \left(\frac{1}{T} \sum_t p(y = c \vert x, w_t)\right) \text{log} \left(\frac{1}{T} \sum_t p(y = c \vert x, w_t)\right)
\end{multline}
with $c$ ranging over all classes and $p(y = c \vert x, w_t)$ being the softmax probability of input $x$ being in class $c$ given the current set of model weights $w_t$. $T$ denotes the number of MCD samples. The Mutual Information (\I) between this posterior and the predictive (softmax) distribution isolates the model uncertainty. It is calculated as follows:
\begin{multline} \label{eq:mutual_info}
	\mathbb{I}[y,w \vert x,D_{train}] = \mathbb{H}[y \vert x, D_{train}] \\
	+ \frac{1}{T} \sum_{c,t} p(y = c \vert x,w_t) \text{log} p(y = c \vert x,w_t)
\end{multline}
Both formulas to approximate \H{} and \I{} are based on \cite{Mukhoti2021a}. To convert these two uncertainty metrics into interpretable probabilities in terms of confidence, we normalized them with their respective theoretical maximum. This is only done for visualization purposes and to achieve comparable estimates. 

\subsection{Confidence Calibration Evaluation Metric} \label{subsec:AUSE}
The design of a novel metric to assess a the uncertainty estimation performance of a semantic segmentation model is driven by the need to have a metric which fully captures all aspects of the confidence calibration. Ideally, a well calibrated model is able to predict high uncertainties when it's predictions are inaccurate and low uncertainties when they are accurate, and vice versa. 

This can be explained with an example: usually, the class \person{} is not well represented in large datasets for the semantic segmentation task, because in traffic situations there are less measurements on people than on e.g. roads or buildings. Furthermore, under natural circumstances the human species has a broad variety of outward appearances and humans can find themselves in plentiful different situations (contrary to cars which have a somewhat predefined appearance and a very restricted set of situations in which they can appear). Now if a model correctly predicts the presence of a person, it is not helpful when it always outputs a high predictive uncertainty because the class \person{} was not well represented within its training. The confidence outputs of every class, no matter how well it is represented in the data, should still be meaningful, not only in terms of incorrect but also in terms of correct classifications. 

The metric we propose is based on the idea of the Area Under the Sparsification Error Curve (AUSE) \cite{Ilg2018}. Contrary to the ECE, which is restricted to measure the expected performance in terms of accuracy, this metric is directly applicable to any given task and thus any appropriate performance metric. By allowing a per-class evaluation, its user is able to catch more systematic uncertainty-related phenomena instead of only local relations. It has been used based on the Brier Score \cite{Gustafsson2020}, which is a proper scoring rule for evaluating probabilistic predictions. Yet, we avoid using the Brier Score as it indirectly encourages overconfident predictions by evaluating the deviation from the one-hot encoding. Instead, we suggest using the mIoU directly, which is the common metric to evaluate semantic segmentation models. 

The idea of sparsification plots is to evaluate the pointwise uncertainty estimation by creating a ranking of the uncertainty values. The pixels with the highest uncertainty are then gradually removed and the performance (i.e. the mIoU for semantic segmentation) is evaluated in the process. If the uncertainty 
measure actually reflects the true performance (which it should), the mIoU should monotonically increase. Ilg et al. \cite{Ilg2018} furthermore suggest a normalization based on the best possible ranking according to the ground truth labels to remove the dependence on the model performance. They refer to it as the oracle curve, which allows a fair comparison between the different classes. The difference between the sparsification and the oracle curve is called Sparsification Error. Finally, the AUSE is defined as the area under the resulting curve. The closer the sparsification curve is to the ground-truth-based oracle curve, the smaller are the AUSE values and the better 
is the respective confidence calibration.

We want to emphasize that through the normalization with the oracle, the AUSE isolates the calibration from the model's classification performance. In some corner cases, when the model is not able to properly learn a class and never predicts it correctly, the respective IoU will constantly be $0.0$. In that case the AUSE will always indicate a perfect confidence calibration, because there is no difference between the oracle and the sparsification curve. Combining the AUSE metric with the sparsification plots and the IoU values provides additional information and insights about the model calibration and prevent incomplete interpretations.

We evaluate this metric on every class (per-class AUSE) for the two different training settings described in Sec. \ref{subsec:training}. Calculating the average of all curves results in a combined metric for the given semantic segmentation model.

%% file: sections/results.tex
\section{Results}
\label{sec:results}
In the following, we present he results from our experiments.

\subsection{Uncertainty Estimation Evaluation with AUSE} \label{subsec:AUSE_eval}

For our experiments, we calculated the sparsification plot and the AUSE across 
the full validation dataset (sequence 8 for the SemanticKITTI dataset \cite{Behley2019}) for both 
models to gain a comprehensive and representative measure.

The AUSE value gives an interpretable quantitative measure of the model's confidence calibration. It gives first insights about the overall uncertainty estimation performance (average value for all classes combined) as well as indicators for the different classes. The lower the value, the better the calibration of the uncertainty estimates. For the model trained with the regular cross-entropy loss function, the values for all three uncertainty estimation methods are given in Tab. \ref{tab:AUSE_XE}. The values indicate a very good confidence calibration for well represented classes like \car{} or \road{}, a good confidence calibration for classes like \trunk{} or \trafficsign{} and a moderate confidence calibration for classes like \motorcycle{} or \person{}. 

% \usepackage{graphicx}
\begin{table}[]
	\centering
	\resizebox{\columnwidth}{!}{%
		\begin{tabular}{lllllllll}
			\toprule class 
			& \multicolumn{2}{l}{IoU} & \multicolumn{2}{l}{predictive}   & \multicolumn{2}{l}{model}       & \multicolumn{2}{l}{(learned) data} \\
			& & & \multicolumn{2}{l}{uncertainty}  & \multicolumn{2}{l}{uncertainty} & \multicolumn{2}{l}{uncertainty}    \\ \midrule
			& XE & wXE & XE  & wXE & XE  & wXE & XE  & wXE \\ \cmidrule(lr){2-3} \cmidrule(lr){4-5} \cmidrule(lr){6-7} \cmidrule(lr){8-9}
			\car{}          & 0.93 & 0.95 & 0.05 & 0.02 & 0.05 & 0.02 & 0.05 & 0.02 \\
			\bicycle{}      & 0.21 & 0.38 & 2.60 & 1.58 & 3.56 & 2.13 & 2.48 & 1.49 \\
			\motorcycle{}   & 0.24 & 0.35 & 2.94 & 2.34 & 3.27 & 2.70 & 2.92 & 2.38 \\
			\truck{}        & 0.51 & 0.66 & 1.19 & 1.10 & 1.24 & 1.06 & 1.18 & 1.10 \\
			\othervehicle{} & 0.36 & 0.45 & 2.53 & 1.69 & 2.75 & 1.88 & 2.56 & 1.68 \\
			\person{}       & 0.29 & 0.56 & 2.90 & 0.88 & 3.70 & 1.03 & 2.80 & 0.86 \\
			\bicyclist{}    & 0.00 & 0.55 & 0.00 & 1.88 & 0.00 & 2.23 & 0.00 & 1.87 \\
			\motorcyclist{} & 0.00 & 0.00 & 0.00 & 0.00 & 0.00 & 0.00 & 0.00 & 0.00 \\
			\road{}         & 0.95 & 0.95 & 0.02 & 0.03 & 0.02 & 0.03 & 0.02 & 0.02 \\
			\parking{}      & 0.45 & 0.41 & 2.97 & 3.08 & 3.02 & 3.06 & 2.94 & 3.03 \\
			\sidewalk{}     & 0.83 & 0.82 & 0.38 & 0.41 & 0.38 & 0.41 & 0.38 & 0.41 \\
			\otherground{}  & 0.02 & 0.06 & 0.89 & 1.62 & 0.92 & 1.80 & 0.88 & 1.57 \\
			\building{}     & 0.90 & 0.88 & 0.08 & 0.16 & 0.08 & 0.15 & 0.08 & 0.16 \\
			\fence{}        & 0.54 & 0.49 & 1.87 & 2.37 & 1.85 & 2.34 & 1.86 & 2.35 \\
			\vegetation{}   & 0.89 & 0.88 & 0.15 & 0.19 & 0.16 & 0.21 & 0.15 & 0.18 \\
			\trunk{}        & 0.63 & 0.60 & 1.26 & 0.79 & 1.62 & 0.92 & 1.29 & 0.77 \\
			\terrain{}      & 0.79 & 0.79 & 0.45 & 0.49 & 0.46 & 0.51 & 0.45 & 0.49 \\
			\pole{}         & 0.59 & 0.54 & 1.39 & 0.96 & 1.46 & 1.03 & 1.37 & 0.95 \\
			\trafficsign{}  & 0.36 & 0.41 & 1.78 & 1.56 & 1.93 & 1.65 & 1.86 & 1.62 \\ \midrule
			all             & 0.50 & 0.56 & 1.23 & 1.12 & 1.39 & 1.22 & 1.23 & 1.11 \\ \bottomrule
		\end{tabular}%
	}
	\caption{AUSE values ($\downarrow$) for all classes in the dataset for different models and different types of uncertainty estimation. XE stands for the model trained with cross-entropy, wXE for the model trained with weighted cross-entropy.}
	\label{tab:AUSE_XE}
\end{table}

\begin{figure}
	\centering
	\subfloat[][Class \car{}] {\includegraphics[width=0.5\columnwidth]{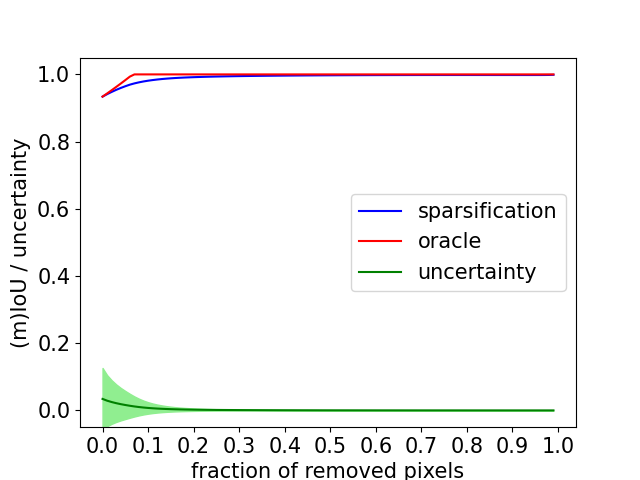}\label{subfig:AUSE_entropy_car_sXE}} 
	\subfloat[][Class \terrain{}] {\includegraphics[width=0.5\columnwidth]{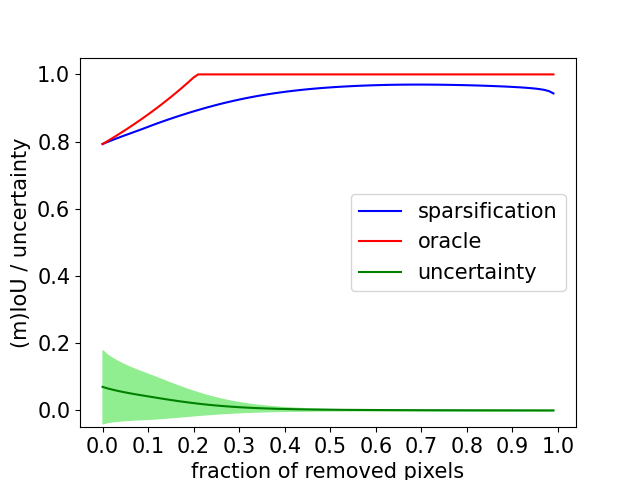}\label{subfig:AUSE_entropy_terrain_sXE}} \\
	\subfloat[][Class \bicycle{}] {\includegraphics[width=0.5\columnwidth]{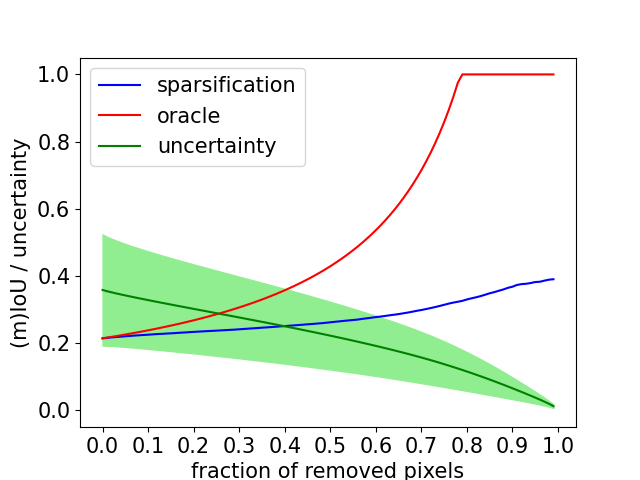}\label{subfig:AUSE_entropy_bycicle_sXE}}
	\subfloat[][Class \bicyclist{}] {\includegraphics[width=0.5\columnwidth]{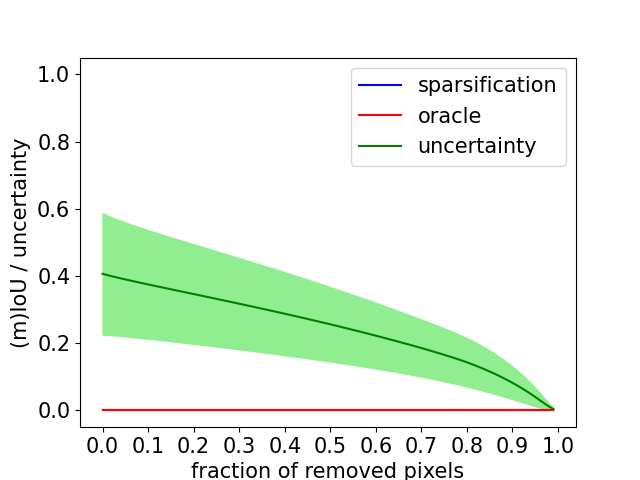}\label{subfig:AUSE_entropy_bicyclist_sXE}}
	\caption{Sparsification plots of predictive entropy for several classes with the cross-entropy-trained model. The mean and standard deviation of the uncertainty is given in green.}
	\label{fig:AUSE_entropy_sXE}
\end{figure}

Noteworthy are the AUSE values for the classes \bicyclist{} and \motorcyclist{} which are exactly $0.0$ for the model trained with regular cross entropy. Here, we can observe the effect described in Sec. \ref{subsec:AUSE}: the IoU for those classes is $0.0$ due to only faulty predictions. This can also be observed in the respective IoU values and sparsification plots (\ref{subfig:AUSE_entropy_bicyclist_sXE}).

The sparsification plots and uncertainty estimates for some exemplary classes are given in Fig. \ref{fig:AUSE_entropy_sXE}. Additionally, the mean and standard deviation of the uncertainty values are given in the plots as well. The y-axis denotes the fraction of the removed points based on their uncertainty. Step by step, the points with the highest uncertainty are removed from the calculations of the mIoU and the mean uncertainty. Resulting, at $y=0.0$ all points and at $y=1.0$ none are included, and at intermediate steps the corresponding percentage. On the x-axis the mIoU is depicted. Overall, the sparsification plot shows the development of the mIoU when gradually removing the points with the highest uncertainty values.

For class \car{} in Fig. \ref{subfig:AUSE_entropy_car_sXE} an almost perfect detection performance and confidence calibration can be observed. The oracle curve shows a large quantity of correctly classified \car{} points (mIoU is almost constantly 1.0) and the sparsification curve follows that trend exactly. The average uncertainty values support the assumption that the class is well learned and properly calibrated.

In Fig. \ref{subfig:AUSE_entropy_terrain_sXE} the sparsification plot for the class \terrain{} is shown. It displays a typical sparsification plot for a well represented class. The oracle curve indicates, that about $80\%$ of the predicted \terrain{} points are correct and the sparsification curve is monotonically increasing. Thereby, uncertainty and its standard deviation is gradually decreasing until it reaches $0.0$. 

Fig. \ref{subfig:AUSE_entropy_bycicle_sXE} shows the sparsification plot of the 
class \bicycle{}, which is not well learned during the training process due to 
its systematic underrepresentation. This is indicated by the sparsification 
curve: it is barely increasing while the oracle curve shows that only about 
$25\%$ of the classifications are correct. Compared to the \terrain{} class, 
the oracle curve is lower and rises later, which indicates a lower IoU for the 
class \bicycle{}. The fact that the sparsification curve is almost constant 
combined with the relatively high uncertainty shows that this class is not well 
represented in the models weights and that there is probably a very high model 
uncertainty (compare Tab. \ref{tab:AUSE_XE}). 

Finally, in Fig. \ref{subfig:AUSE_entropy_bicyclist_sXE} we observe a 
border-case for the sparsification plots and the AUSE metric. From the IoU and 
oracle curve we can conclude that for this particular validation set the model 
was not able to make any correct classification. This imposes a problem: no 
matter which point will be removed first for the plot, the resulting AUSE will 
always be $0.0$. Although, this anomaly points out that there is a general 
problem with this class in particular. From this plot we can conclude the model 
had problems to learn typical \bicyclist{} features, which might be due to 
either an underrerpresentation or an incoherent representation of the data, or 
both. Also it could indicate that the architecture or training procedure needs 
optimization.

\subsection{Model Training Adaption} \label{subsec:weightedXE}

\begin{figure}
	\centering
	\subfloat[][Cross-entropy] {\includegraphics[width=0.5\columnwidth]{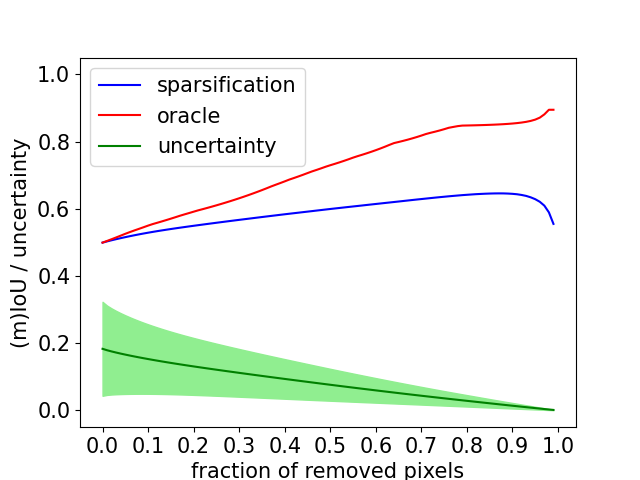}\label{subfig:AUSE_entropy_allclasses_sXE}} 
	\subfloat[][Weighted cross-entropy] {\includegraphics[width=0.5\columnwidth]{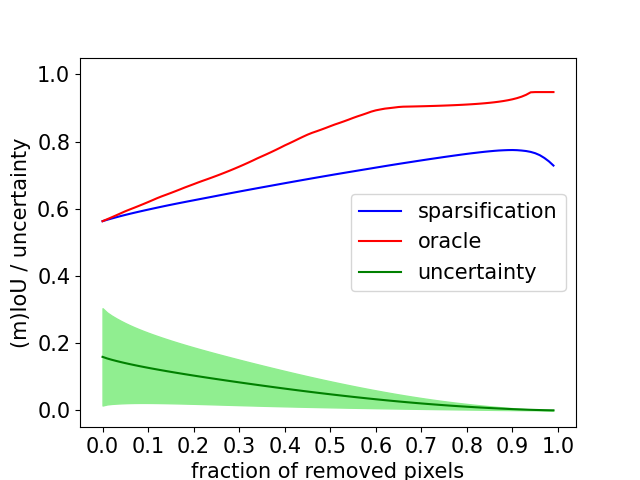}\label{subfig:AUSE_entropy_allclasses_wsXE}} 
	\caption{Sparsification plots of the predictive entropy over all classes of two models trained with the cross-entropy loss and the weighted cross-entropy loss, respectively. The mean and standard deviation of the uncertainty is given in green.}
	\label{fig:AUSE_comparsion_sXEwsXE}
\end{figure}

The above analysis reveals a bad confidence calibration for classes which are generally underrepresented in the data and thus not well learned by the model. To verify this assumption, we conducted the same evaluations on a model which is trained with a weighted loss: the cross-entropy loss is weighted with the respective classes inverse log-frequency within the dataset. We expected a reduced model uncertainty about the underrepresented classes and at the same time a better confidence calibration of them.

When comparing the overall AUSE and sparsification plots (Fig. \ref{fig:AUSE_comparsion_sXEwsXE}) for the predictive uncertainty, a major improvement of the confidence calibration can be observed. As can be seen in Tab. \ref{tab:AUSE_XE}, the AUSE for several underrepresented classes improved drastically, though at the expense of other classes which performed well before (e.g. \parking{}, \fence{}). 

\begin{figure}
	\centering
	\subfloat[][Class \car{}] {\includegraphics[width=0.5\columnwidth]{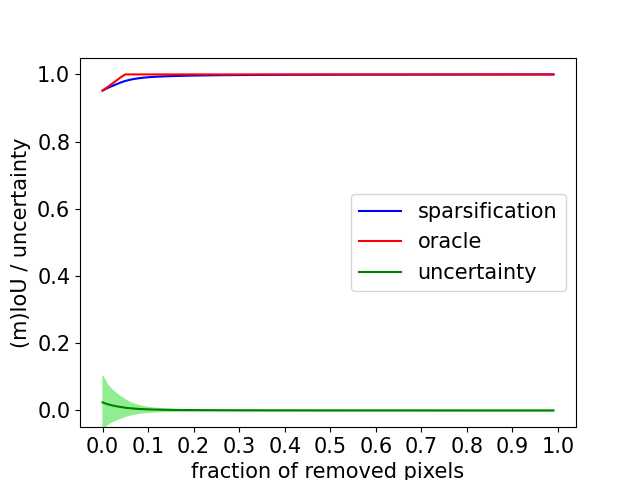}\label{subfig:AUSE_entropy_car_wsXE}} 
	\subfloat[][Class \terrain{}] {\includegraphics[width=0.5\columnwidth]{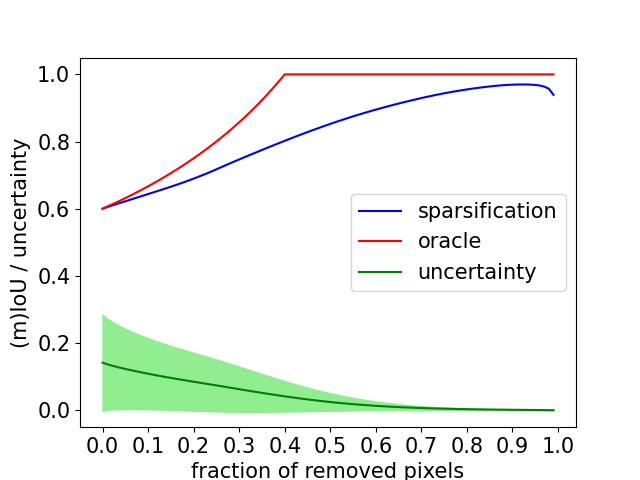}\label{subfig:AUSE_entropy_terrain_wsXE}} \\
	\subfloat[][Class \bicycle{}] {\includegraphics[width=0.5\columnwidth]{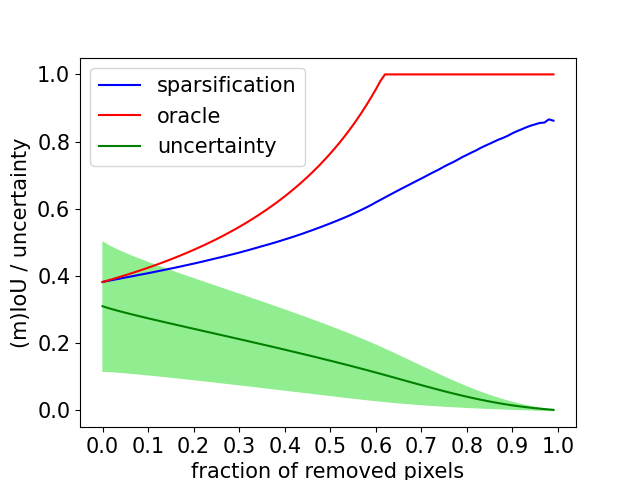}\label{subfig:AUSE_entropy_bycicle_wsXE}}
	\subfloat[][Class \bicyclist{}] {\includegraphics[width=0.5\columnwidth]{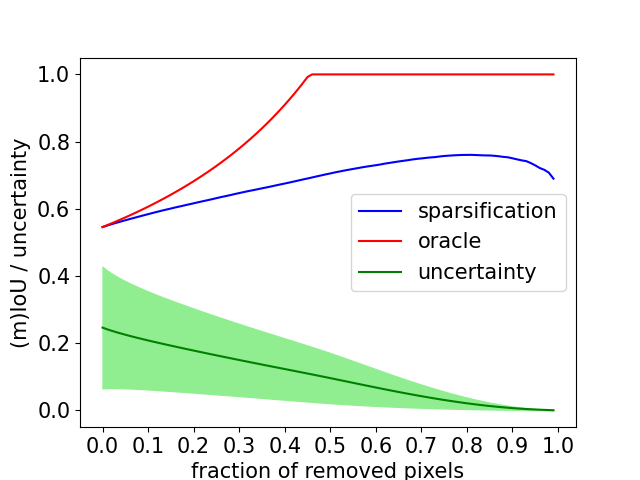}\label{subfig:AUSE_entropy_bicyclist_wsXE}}
	\caption{Sparsification plots of predictive entropy for several classes with the weighted cross-entropy-trained model. The mean and standard deviation of the uncertainty is given in green.}
	\label{fig:AUSE_entropy_wsXE}
\end{figure}

As Fig. \ref{fig:AUSE_entropy_wsXE} reveals, previously well calibrated classes 
like \car{} (Fig. \ref{subfig:AUSE_entropy_car_wsXE}) still exhibits a 
reasonable uncertainty estimation. Classes like \terrain{} went through a 
slight decline of the calibration performance, likely due to the confusion with 
some underrepresented classes which were given more weight in the calculation 
of the loss during the training. In the sparsification plot in Fig. 
\ref{subfig:AUSE_entropy_terrain_wsXE} this is reflected in the curves and can 
also be seen in the slope of the mean uncertainty of the \terrain{} class 
points.

For the case of the \bicyclist{} class (Fig. 
\ref{subfig:AUSE_entropy_bicyclist_wsXE}), the AUSE increased but solely 
because the oracle curve is not constantly at an IoU of $0.0$ (compare Tab. 
\ref{tab:AUSE_XE}). The previously poor performing classes like \bicyclist{} 
and even \bicycle{} experienced a major detection improvement (IoU) through the 
change of the loss function and thus a reduction of the (model) uncertainty, as 
can be seen in Fig. \ref{subfig:AUSE_entropy_bycicle_wsXE}. These insights can 
be used to balance this tradeoff with respect to the desired model performance, 
e.g. to carefully tune the detection of more safety-critical classes in the 
context of autonomous driving. 

\subsection{Detection and Analysis of Label Problems} \label{subsec:label_problems}

For datasets with given labels, a mask can be created which filters only incorrect classifications accompanied by a 
low uncertainty estimate. This allows us to quickly identify potential label issues which can be further analyzed.

\subsubsection{Problems Concerning the Label Hierarchy}
\begin{figure}
	\centering
	\subfloat[][Ground truth labels] {\includegraphics[width=0.9\columnwidth]{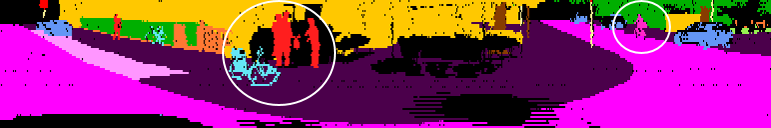}\label{subfig:bycicle_bicyclist_confusion_1900gt}} \\
	\subfloat[][Model predictions] {\includegraphics[width=0.9\columnwidth]{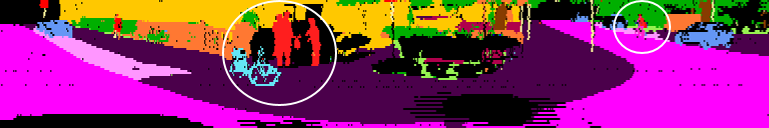}\label{subfig:bycicle_bicyclist_confusion_1900pred}} \\
	\subfloat[][Predictive uncertainty] {\includegraphics[width=0.9\columnwidth]{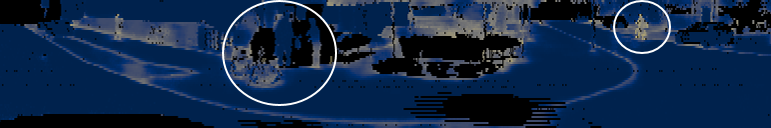}\label{subfig:bycicle_bicyclist_confusion_1900pred_u}} \\
	{\scriptsize semantic segmentation: \textd{road} \textd{sidewalk} \textb{parking} \textb{building} \textd{vegetation} \textd{car} \textd{person} \textd{fence} \textb{bicycle} \textd{trunk} \textd{bicyclist} \textb{pole} \textb{terrain} \textd{otherground} \\
	uncertainty estimation: \textd{low} \textb{high}} \\
	\caption{Example of a confusion of the classes \bicycle{} / \person{} and \bicyclist{} (right circle, compared to left circle). Instead of recognizing the \bicyclist{} as one entity, the model predicts the rider as \person{} and the bicycle as \bicyclist{}, both with high predictive uncertainties.}
	\label{fig:confusion_bicycle_bicyclist_1900}
\end{figure}

\begin{figure}
	\centering
	\subfloat[][Ground truth labels] {\includegraphics[width=0.9\columnwidth]{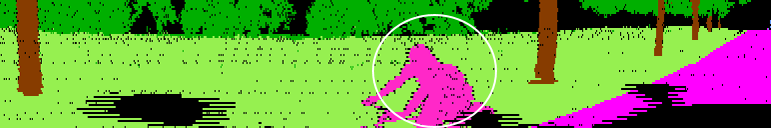}\label{subfig:bycicle_bicyclist_confusion_3050gt}} \\
	\subfloat[][Model predictions] {\includegraphics[width=0.9\columnwidth]{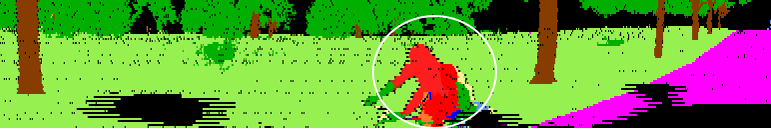}\label{subfig:bycicle_bicyclist_confusion_3050pred}} \\
	\subfloat[][Predictive uncertainty] {\includegraphics[width=0.9\columnwidth]{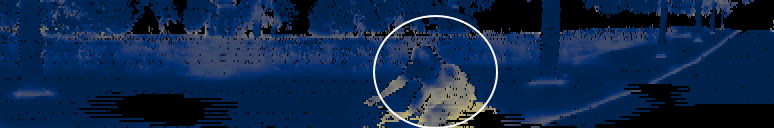}\label{subfig:bycicle_bicyclist_confusion_3050pred_u}} \\
	{\scriptsize semantic segmentation: \textd{vegetation} \textb{terrain} \textd{trunk} \textd{road} \textd{bicyclist} \textd{person} \textb{pole} \textd{motorcycle} \textb{bicycle} \\
	uncertainty estimation: \textd{low} \textb{high}} \\
	\caption{Example of a confusion of the classes \bicycle{} / \person{} and \bicyclist{}. Although the person is definitively riding a bicycle, the model predicts the rider as \person{} with a high confidence.}
	\label{fig:confusion_bicycle_bicyclist_3050}
\end{figure}

The first case can be observed when closely analyzing the detection and 
calibration performance of the class \bicyclist{}. Despite the detection 
improvement of the class \bicyclist{} a little drop of the sparsification curve 
can be observed in Fig. \ref{subfig:AUSE_entropy_bicyclist_wsXE}. This 
indicates that the model made some wrong predictions (either false positives or 
false negatives) of the class \bicyclist{} but was at the same time quite 
confident about its prediction (low uncertainty). Further investigations of all 
the models' predictions concerning the class \bicyclist{} reveal that indeed 
the model had problems with the label hierarchy in this case. Fig. 
\ref{fig:confusion_bicycle_bicyclist_1900} shows a projected LiDAR scan that 
illustrates this phenomenon. In the left white circle, we find representatives 
of the classes \bicycle{} and \person{}, in the right one a \bicyclist{}. 
Nevertheless, since CNNs primarily learn typical local class features, it is 
understandable when the model predicts a \bicycle{} and a \person{} separately. 
While in this frame the uncertainty for the two examples are (correctly) high, 
another example demonstrates the problem resulting from this way of labeling. 
Fig. \ref{fig:confusion_bicycle_bicyclist_3050} shows a rider on a bicycle, 
labeled as \bicyclist{} (\ref{subfig:bycicle_bicyclist_confusion_3050gt}), 
which he clearly is but the bicycle is not fully visible. The model predicts a 
\person{} with a very high confidence, which can be seen Fig. 
\ref{subfig:bycicle_bicyclist_confusion_3050pred} and 
\ref{subfig:bycicle_bicyclist_confusion_3050pred_u}. These kind of point 
clusters lead to a drop in the sparsification curve and thus to a worse 
confidence calibration, despite the rider on the bicycle is also a \person{} without any 
doubt. 

\subsubsection{Problems concerning the label}
\begin{figure}
	\centering
	\subfloat[][Ground truth labels] {\includegraphics[width=0.9\columnwidth]{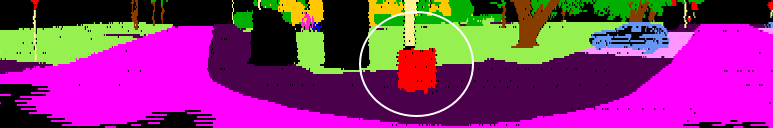}\label{subfig:sign_pole_confusion_2900gt}} \\
	\subfloat[][Model predictions] {\includegraphics[width=0.9\columnwidth]{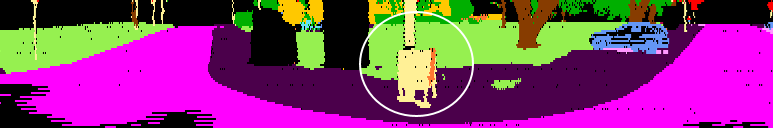}\label{subfig:sign_pole_confusion_2900pred}} \\
	\subfloat[][Predictive uncertainty] 
	{\includegraphics[width=0.9\columnwidth]{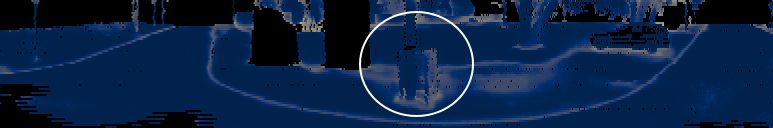}\label{subfig:sign_pole_confusion_2900pred_u}}
	 \\
	\subfloat[][Camera image] 
	{\includegraphics[width=0.9\columnwidth]{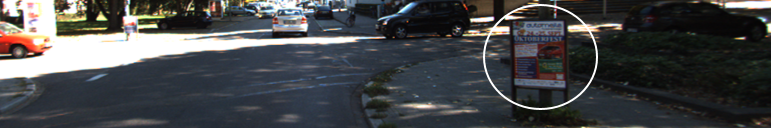}\label{subfig:sign_pole_confusion_2900png}}
	 \\
	{\scriptsize semantic segmentation: \textd{road} \textb{terrain} \textd{sidewalk} \textb{pole} \textd{trafficsign} \textd{trunk} \textd{vegetation} \textb{building} \textd{bicyclist} \textd{motorcycle} \textb{parking} \textd{car} \textd{fence}\\
	uncertainty estimation: \textd{low} \textb{high}} \\
	\caption{Example of a wrongly labeled \trafficsign{}. The sign in question is a advertising poster, which can be seen in the camera footage. The model erroneously predicts one side as part of the pole behind it with a low uncertainty.}
	\label{fig:sign_pole_confusion2900}
\end{figure}

\begin{figure}
	\centering
	\subfloat[][Ground truth labels] {\includegraphics[width=0.9\columnwidth]{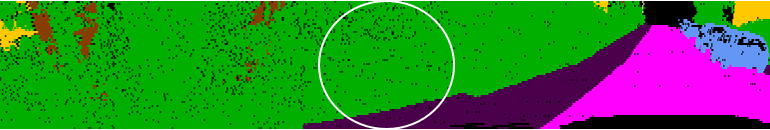}\label{subfig:fence_vegetation_confusion_100gt}} \\
	\subfloat[][Model predictions] {\includegraphics[width=0.9\columnwidth]{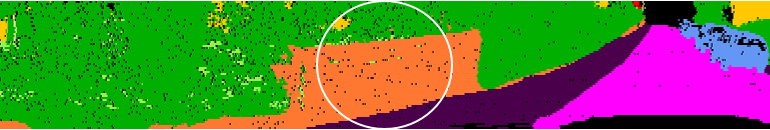}\label{subfig:fence_vegetation_confusion_100pred}} \\
	\subfloat[][Predictive uncertainty] 
	{\includegraphics[width=0.9\columnwidth]{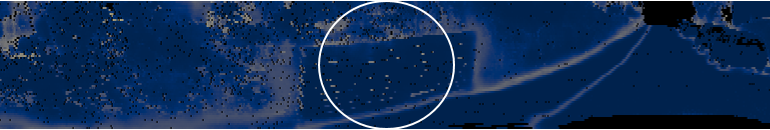}\label{subfig:fence_vegetation_confusion_100pred_u}}
	 \\
	\subfloat[][Camera image] 
	{\includegraphics[width=0.9\columnwidth]{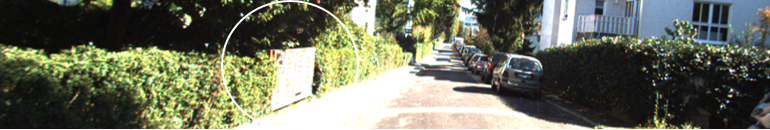}\label{subfig:fence_vegetation_confusion_100png}}
	 \\
	{\scriptsize semantic segmentation: \textd{vegetation} \textd{trunk} \textd{sidewalk} \textd{road} \textd{car} \textb{building} \textb{vegetation} \textd{fence} \textd{trafficsign} \\
	uncertainty estimation: \textd{low} \textb{high}} \\
	\caption{Example of a gate wrongly labeled as \vegetation{}. The model 
	recognizes a \fence{} with high confidence, which is backed up by the 
	uncertainty image and the camera footage.}
	\label{fig:fence_vegetation_confusion_100}
\end{figure}

For the second case we found examples of classes which where inconsistently 
defined. Two examples can be seen in Fig. \ref{fig:sign_pole_confusion2900} and 
Fig. \ref{fig:fence_vegetation_confusion_100}. The first one shows a generic 
sign that is wrongly labeled as \trafficsign{} (Fig. 
\ref{subfig:sign_pole_confusion_2900gt}), the model assumes it partially 
belongs to the pole behind it (Fig. \ref{subfig:sign_pole_confusion_2900pred}), 
in lack of a more suitable class. In the second figure the model detects a 
fence embedded in a hedge (Fig. 
\ref{subfig:fence_vegetation_confusion_100pred}). This fence is not labeled 
accordingly, instead it is incorrectly labeled as part of the hedge (Fig. 
\ref{subfig:fence_vegetation_confusion_100gt}). In both cases, the models' 
predictive uncertainty is low (Fig. \ref{subfig:sign_pole_confusion_2900pred_u} 
and Fig. \ref{subfig:fence_vegetation_confusion_100pred_u}), indicating that 
there is probably a problem with the provided labels in this region. In these 
cases, a further inspection helps to identify and mitigate label errors.

%% file: sections/discussion.tex
\section{Discussion}
\label{sec:discussion}

Our analyses demonstrated how our proposed metric allows a qualitative and quantitative evaluation of the confidence calibration of a semantic segmentation model. 
It has several advantages compared to existing confidence calibration metrics for semantic segmentation models: 
\begin{itemize}
	\item Since it is based on the mIoU metric, it does not neglect underrepresented classes and allows an independent confidence calibration evaluation of all classes.
	\item It does not require any parameter tuning (like accuracy and uncertainty thresholds of the PAvPU metric).
	\item It does not depend on a specific normalization method of the uncertainty estimates (which is necessary for many uncertainty estimation methods).
\end{itemize}

%Combined with the sparsification plots the AUSE provides some valuable 
%insights 
%about the training process and helps to detect problems within the underlying 
%data. It helps to fine-tune the model training, e.g. to chose a loss function 
%or an architecture which can handle underrepresented data. Furthermore it 
%gives 
%hints about problems with the labels and helps to uncover ambiguous label 
%hierarchies or label errors. We demonstrate this by using the predictive 
%uncertainty as an indicator how much an entity resembles its typical 
%appearance, points with questionable or problematic labels can automatically 
%be 
%found. Once detected, these issues can be resolved by e.g. correcting or 
%re-defining the labels so that they better represent real-world concepts.

From the sparsification plots we know that our model made wrong predictions 
with low uncertainties. These highly confident but wrong predictions manifest 
in more or less grave drops in the sparsification curve. Ideally, a well 
calibrated model would exhibit strictly increasing curves, indicating that 
the model meets all previously mentioned (compare Sec. \ref{subsec:u_metric}) 
properties:
\begin{itemize}
	\item accuracy in case of confident predictions
	\item uncertainty in case of inaccurate predictions
	\item confidence in case of accurate predictions
\end{itemize}
There are three possible reasons for highly confident but wrong predictions: 1. 
the situation is difficult to interpret and the models performance is 
insufficient, 2. there are problems with the label hierarchy or definition, 3. 
there are problems with the labels. 

In the former situation, the problem does not lie within the labels but rather 
the model performance. In this case, it is worth analyzing whether the failure 
case is critical and how to mitigate it. We demonstrated one example measure in 
Sec. \ref{subsec:weightedXE}: adjusting the loss function in favor of 
underrepresented classes. This can be further adjusted and refined until the 
model performance meets the requirements. Nevertheless, the other two cases 
cannot be eliminated by adjusting the model but by questioning the training 
data (as elaborated in Sec. \ref{subsec:label_problems}).

The second case describes conceptual problems with the labels and their 
definitions. When there is a problem with the label hierarchy, the model 
is not able to distinguish between two usually seperated components and the new 
entity they built when combined. Inconststent labeling in ambiguous situations 
also falls into this category. Examples for both cases can be observed in Fig. 
\ref{fig:example_motorcyclist}. The model detects the rider as a \person{} and 
the vehicle as a \motorcycle{}, which is not wrong. Despite not riding on a 
road, the labels define it as one entity, a \motorcyclist{}. Furthermore, the 
wall on the left side is labeled as fence but the model predicts it as part of 
the building instead. In both cases it is debatable which definition is closer 
to the real situation and, above all, which definition is desireable and 
suitable for downstream tasks.

The third case describes label errors which might be due to weaknesses in the 
labeling process or human mistakes. Once found they can easily be corrected 
accordingly.

After identifying problematic classes with the help of the AUSE and 
sparsification curves, measures can be taken to eliminate the three reasons as 
far as possible.

In the rare case that the model is not able to learn a representation of a class, the AUSE reaches its limits. As mentioned in Sec. \ref{subsec:AUSE_eval}, the AUSE alone is hard to evaluate. Nevertheless, in this case the respective IoU and the sparsification plots help to detect and alleviate the underlying problem.

%% file: sections/conclusion.tex
\section{Conclusion}
\label{sec:conclusion}

In this paper we proposed a novel calibration metric for point-wise multi-class classification models. We demonstrated its capabilities in finding limitations in the training process and problems in the label structure. By closely analyzing the results we were able to point out unclear and problematic class definitions and label inconsistencies.

This illustrates how crucial the quality of the database is for the model training and uncertainty estimation, especially when it comes to imbalanced data. When evaluating a models' calibration, we have to bear in mind the way how the labeling effects the models uncertainty estimation. Thus, we have to be careful with the way, concepts and classes are represented in the labels and the label hierarchy.

Concluding, our analyses yielded important insights about uncertainty 
estimation in semantic segmentation models, their calibration and the role the data representation plays.

To further advance in this field of research, it is important to entangle the different types of uncertainty and gain a deeper understanding of their role in model training. This might provide access to identifying and alleviating limiting factors within the process, such as data underrepresentation or inconsistent class definitions and might even have an impact on related tasks like domain adaptation or outlier detection.